\documentclass{article}
\usepackage{spconf, amsmath, amssymb, graphicx}
\usepackage{array}
\usepackage{setspace}
\usepackage{amsmath}
\usepackage{todonotes}

\usepackage{enumitem}
\usepackage[numbers,sort&compress]{natbib}
\setlist{nosep, leftmargin=14pt}

\usepackage{mwe} 

\title{Follow Your Heart: Landmark-Guided Transducer Pose Scoring for Point-of-Care Echocardiography}
\name{
\begin{minipage}{\textwidth}
    \centering
    Zaiyang Guo$^{\ddagger \star}$, Jessie N. Dong$^{\ddagger \star}$, Filippos Bellos$^{\dagger}$,
    Jilei Hao$^{\star}$, Emily J. MacKay$^{\star}$, Trevor Chan$^{\star}$,
    Shir Goldfinger$^{\star}$, Sethu Reddy$^{\S}$, Steven Vance$^{\S}$,
    Jason J. Corso$^{\dagger}$, Alison M. Pouch$^{\star}$%
\end{minipage}
\thanks{$^{\ddagger}$ These authors contributed equally.\\
This research was funded by ARPA-H grant 1AY2AX000062.}
}
\address{
{$^{\star}$ University of Pennsylvania, Philadelphia, PA, USA} \\
$^{\dagger}$ University of Michigan, Ann Anbor, MI, USA \\
$^{\S}$ Central Michigan University, Mt Pleasant, MI, USA
}
\begin{document}

%
\maketitle
\begin{abstract}
Point-of-care transthoracic echocardiography (TTE) makes it possible to assess a patient’s cardiac function in almost any setting. A critical step in the TTE exam is acquisition of the apical 4-chamber (A4CH) view, which is used to evaluate clinically impactful measurements such as left ventricular ejection fraction (LVEF). However, optimizing transducer pose for high-quality image acquisition and subsequent measurement is a challenging task, particularly for novice users. In this work, we present a multi-task network that provides feedback cues for A4CH view acquisition and automatically estimates LVEF in high-quality A4CH images. The network cascades a transducer pose scoring module and an uncertainty-aware LV landmark detector with automated LVEF estimation. A strength is that network training and inference do not require cumbersome or costly setups for transducer position tracking. We evaluate performance on point-of-care TTE data acquired with a spatially dense ``sweep" protocol around the optimal A4CH view. The results demonstrate the network’s ability to determine when the transducer pose is on target, close to target, or far from target based on the images alone, while generating visual landmark cues that guide anatomical interpretation and orientation. In conclusion, we demonstrate a promising strategy to provide guidance for A4CH view acquisition, which may be useful when deploying point-of-care TTE in limited resource settings.
\end{abstract}
\section{Introduction}
\label{sec:intro}

Transthoracic echocardiography (TTE) is the first-line imaging modality for the assessment of cardiac function due to its portability, lack of ionizing radiation, and ease of rapid acquisition \cite{cheitlin_accaha_1997}. In the last decade, advancement in point-of-care TTE platforms---comprising a handheld transducer connected to a smartphone or tablet---have made it possible to carry out this exam almost anywhere, including resource-limited communities and physically restricted settings \cite{kornelsen_rural_2023}. One of the critical functional measurements TTE can provide is left ventricular ejection fraction (LVEF), the percentage of total blood in the left ventricle that is ejected to the rest of the body with each heartbeat. It is a key quantitative metric of global left ventricular systolic function in a multitude of scenarios when patients have electrocardiographic abnormalities, palpitations, stroke, peripheral embolic events, syncope, lightheadedness, chest pain, or arrhythmia \cite{american_college_of_cardiology_foundation_appropriate_use_criteria_task_force_accfaseaha_2011}. 

During a TTE exam, the clinician places the transducer on the patient's chest in various positions to acquire cross-sectional images of the heart. A standard TTE exam has $>$20 standard views through different cross-sections \cite{mitchell_guidelines_2019}; we focus on the apical 4-chamber (A4CH) view, which captures both atria and ventricles and is commonly used for reproducible LVEF assessment. Optimizing transducer pose for high-quality view acquisition is highly user dependent and is challenging for novice users or medical providers who do not perform echocardiography frequently \cite{bick_comparison_2013}.

Prior work has aimed to provide A4CH view guidance based on transducer tracking with additional equipment. Bao et al.~\cite{linguraru_real-world_2024} use an electromagnetic tracking system to track transducer rotations while acquiring TTE images and train a two-stage model: a feature extractor to predict rotation between two frames and a second guidance module to provide rotation instruction. Jiang et al.~\cite{jiang_cardiac_2024} adopted a comparable approach, employing a robotic ultrasound system to record the transducer pose. Similarly, Milletari et al.~\cite{milletari_straight_2019} and Li et al.~\cite{li_autonomous_2021} utilized reinforcement-learning frameworks coupled with optical tracking and a robotic manipulator, respectively. Soemantoro et al.~\cite{soemantoro_ai-powered_2023} use an camera to track the transducer, and incorporate multiple deep learning models for image quality assessment, transducer tracking and finding standard view windows. Pasdeloup et al.~\cite{pasdeloup_real-time_2023} contribute a pilot study in which view guidance is simulated by slicing 2D frames from 3D cardiac echo images to emulate the images obtained due to translation and rotation from the optimal transducer pose, and train rotation and tilt networks to predict guidance signals. Commercial software offering interactive TTE acquisition guidance has recently emerged~\cite{trost_artificial_2025}, yet the proprietary nature of these systems limits information about their underlying architectures and training paradigms, precluding direct methodological comparison with other approaches.

Although these works have clearly demonstrated the potential for interactive guidance in A4CH view acquisition, most of them rely on costly and cumbersome setups for transducer tracking. In this work, we require no such setup and provide guidance directly from echocardiographic images. Our network-based approach guides A4CH view acquisition and automatically estimates LVEF in high-quality A4CH images. All data were collected with institutional approval and informed consent from human subjects. The key contributions of this work are:

\begin{enumerate}
    \item A TTE acquisition protocol that enables training of a transducer score model with minimal setup and patient data.
    \item A multi-task network that provides visual landmark cues for the cardiac chambers, scores the quality of transducer pose for the A4CH view into "on target" (green), "close to target" (yellow), or "far from target" (red) categories, and streamlines LVEF estimation in hiqh-quality A4CH images.
    \item Verification of the multi-task network in datasets acquired on human subjects with point-of-care TTE.
\end{enumerate}

\begin{figure}
    \centering
    \includegraphics[width=1\linewidth]{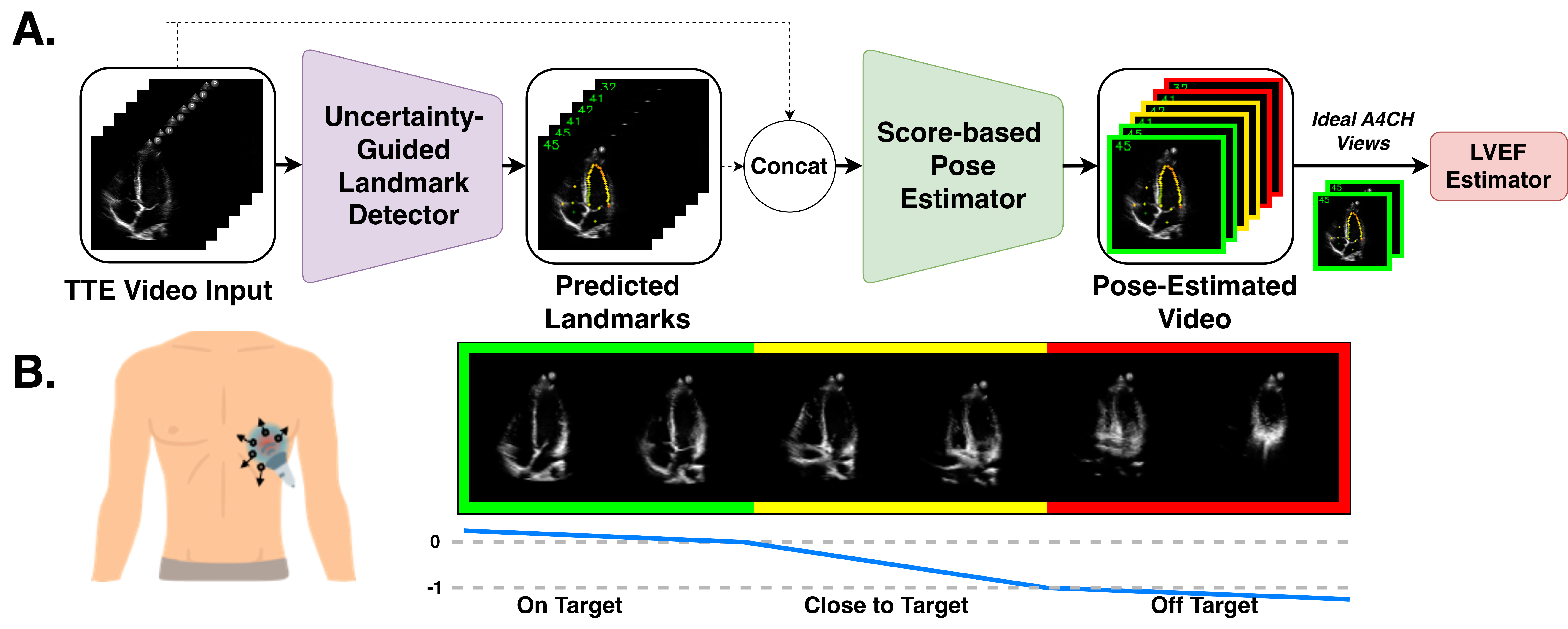}
    \caption{\textbf{A.} Overview of the multi-task network. TTE video frames are first passed through the landmark detector. The images and predicted landmarks then pass through the pose scoring module to be classified as green, yellow, or red. Optimal "green" clips are then passed to an automated LVEF estimator. \textbf{B.} Each TTE sweep starts with the transducer in the optimal A4CH pose and drifts away from the optimal pose. Ground truth pose categories (green, yellow, red) are manually assigned and mapped to a numeric pose score.}
    \label{fig:pipeline}
\end{figure}

\section{Methods}
\subsection{Multi-Task Network Overview}
An overview of the multi-task network for A4CH view guidance and LVEF estimation is shown in Fig.~\ref{fig:pipeline}A. Briefly, the network cascades an uncertainty-aware landmark detector with a transducer pose scoring module. The intuition is that the transducer position and orientation are optimal when the target anatomy (specified by landmarks) is within the field of view. Images that are scored as diagnostic-quality A4CH views are subsequently passed to an LVEF estimator. The scoring system and each network component are described in the following sections.

\subsection{Uncertainty Aware Landmark Detection}
\begin{figure}
    \centering
    \includegraphics[width=1\linewidth]{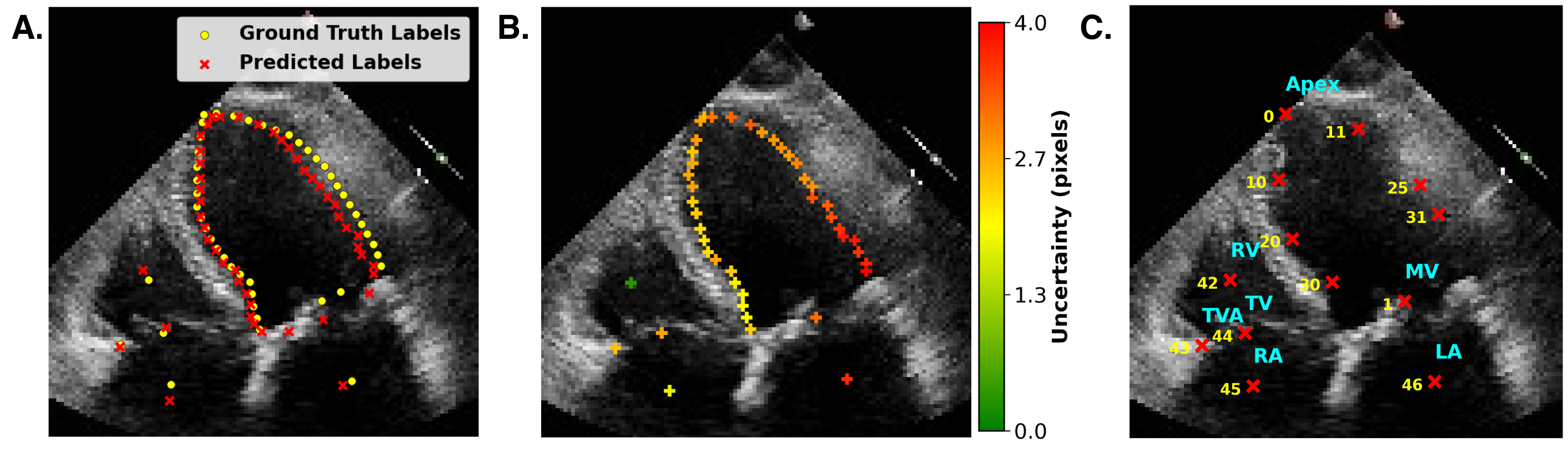}
    \caption{\textbf{A.} Predicted and ground truth landmarks \textbf{B.} Predicted landmarks displayed with prediction uncertainty \textbf{C.} Key landmark output passed on to pose scoring module}
    \label{fig:landmark_result}
\end{figure}

The landmark detection module leverages the EchoNet Dynamic dataset~\cite{ouyang_video-based_2020}, which comprises 10,000 video clips, each with manual annotations on one diastolic and one systolic frame, totaling 20,000 annotated frames. Each frame includes 42 landmarks that define the LV contour. In addition, we annotate a subset of approximately 3,000 frames with 5 additional landmarks: the right ventricle (RV), right atrium (RA), left atrium (LA), mid-plane of the tricuspid valve (TV), and the tricuspid valve annulus (TVA). For each frame, we specify a landmark visibility score: 1 - high confidence, 2 - moderate confidence, 3 - low confidence. In total, the module identifies 47 landmarks capturing key structures in the A4CH view (Fig.\ref{fig:landmark_result}A,C). 

We formulate the landmark detection problem as a semantic segmentation task. Specifically, we employ a modified ResNet34 architecture that uses pre-trained weights on ImageNet \cite{he_deep_2016}. We utilize the initial ResNet34 layers as our encoder, followed by a progressively larger decoder designed to capture fine-grained landmark information \cite{mccouat_contour-hugging_2022}. The decoder contains five upsampling layers (channel dimension: 512, 256, 128, 64, and 64), each block contains a batch normalization layer and a ReLU activation. Following the final decoder layer, we use a $1 \times 1$ convolution layer to map feature maps into 47 channels. To train the network to handle partially available landmarks and varying visibility levels, we use a weighted negative log likelihood loss.


To compute this loss, we reshape logits from $(B, L, H, W)$ to $(B, L, H \times W)$ and apply \texttt{log\_softmax} over the spatial dimension. We convert each ground-truth landmark coordinate $(x, y)$ into a 1D index ($y \times W + x$), clamp it to valid bounds, and extract the corresponding log-probabilities. A boolean mask $(B, L)$ identifies annotated landmarks, and we weight their log-probabilities by this mask and the sample-wise visibility weights $\text{vis\_w} \in \mathbb{R}^{B}$, thereby emphasizing highly visible landmarks~\cite{kumar_luvli_2020}. The batch loss function for batch size $B$ is:

\footnotesize
\begin{align}
\mathcal{L} = \frac{1}{B} \sum_{i=1}^{B} \left( -\sum_{j=1}^{L} \log \left( \frac{\exp(\text{logits}_{ij})}{\sum_{k} \exp(\text{logits}_{ik})} \right) \cdot \text{mask}_{ij} \cdot \text{vis\_w}_i \right)
\end{align}
\normalsize
%
Finally, to extract uncertainty, we apply a 2D softmax to each landmark channel, yielding spatial probability distributions. The uncertainty is reported as the radius of non-zero regions in these distributions~\cite{mccouat_contour-hugging_2022}.

The EchoNet Dynamic dataset is partitioned with an 80\%/10\%/10\% train/validation/test split. We fine-tune the pre-trained ResNet34 encoder for 20 epochs using the Adam optimizer at a learning rate of 0.001, with a batch size of 512.

\subsection{Transducer Pose Scoring Module}
\label{pose_scoring_method}  We use a ``traffic light" metaphor to capture transducer pose quality scoring: on target (green), close to target (yellow), or far from target (red) relative to the ideal A4CH view. Training the module involves systematic image acquisition illustrated in Fig.~\ref{fig:pipeline}B. TTE from each patient is acquired in a series of approximately 40 ``sweeps." Each sweep is a 10-second video clip wherein the expert  sonographer starts with the transducer in the optimal A4CH pose and gradually moves away in a random trajectory. Altogether, the sweeps apply random pose transformations relative to the optimal A4CH view that are confined to and capture the entire anatomical region illustrated in Fig.~\ref{fig:pipeline}B. Transducer pose scores (green, yellow, red) are collaboratively assigned to image frames within each sweep by two individuals based on the point-based criteria described in Table \ref{tab:echo_scoring}. 

We investigate two approaches for pose scoring. The first approach uses ResNet18 and formulates the problem as a regression task to capture the continuous nature of pose changes. Green, yellow, and red classes are each mapped to a continuous pose score as shown in Fig.\ref{fig:pipeline}B. Green scores range from 1 to 0, yellow from 0 to -1, and red from -1 to -2. The score assigned to each frame is subsequently used as the ground truth score for the regression task. We conduct subject-level 5-fold cross validation in which each fold uses all frames from two subjects as the test set, one subject as the validation, and the rest as training set.  This module is trained for 100 epochs with the Adam optimizer and a class-weighted mean squared error loss function, and the weights with the lowest average validation loss of the last 5 epochs are saved. Augmentation including brightness and contrast, uniform scaling, horizontal and vertical translation are applied during training. 

The second approach explores a multi-modal LLM architecture (LLaMA \cite{touvron_llama_2023}) conditioned on both image data and task instructions for transducer pose scoring. Following LLaMA-Adapter \cite{zhang_llama-adapter_2024}, we integrate visual information from TTE frames into the LLaMA model by combining encoded image features with adaption prompts. The frames are first processed through a ResNet18 encoder to extract visual features, which are then projected into the LLM embedding space. Task instructions are tokenized and mapped to text embeddings via the frozen LLaMA embedding layer before being passed as input to the language model. To generate the final prediction, a projection layer takes the embeddings from the last hidden layer of the LLM and maps them to the three pose categories. This approach maintains the LLaMA model parameters frozen while only training the ResNet encoder, the projection layers, and the adaption prompts. This approach is trained for 100 epochs using the AdamW optimizer, and a cyclical learning rate scheduler. The same augmentations are applied.

\subsection{Fully Automated LVEF Estimation}

We fine-tune a deep R(2+1)D model \cite{tran_closer_2018} with EchoNet Dynamic training data and code \cite{ouyang_video-based_2020} for  100 epochs with default train-test split in EchoNet. Video segments that have at least 26 frames or 1 second duration are passed to the model. 

\begin{table}[h]
    \centering
    \resizebox{0.49\textwidth}{!}{
    \begin{tabular}{l c}
        \hline
        \textbf{Criterion} & \textbf{Deduction (points)} \\
        \hline
        LV free wall not visible & -1 \\
        RV free wall not visible & -0.5 \\
        LA entirely/partially out of view & -2 / -1 \\
        RA not visible & -0.5 \\
        Aorta clearly visible (apical 5-chamber view) & -1 \\
        Other significant signal dropout & -0.5 \\
        \hline
 \multicolumn{2}{c}{\textbf{Red} if total deduction $\geq$ 2, else \textbf{yellow} if total deduction $\geq$ 1, else \textbf{green}.}\\
\hline
    \end{tabular}
    }
    \caption{Pose scoring criteria for TTE frames. }
    \label{tab:echo_scoring}
\end{table}

\section{Results}
The TTE acquisition protocol (Section~\ref{pose_scoring_method}) was carried out on nine individuals, three male and six female, using the Clarius PAL HD3 Wireless Ultrasound Scanner (Clarius Mobile Health) and the Lumify handheld ultrasound imaging system (Philips North America). We collected 88,524 images with 430 sweeps. All images were manually scored according to the criteria described in Section \ref{pose_scoring_method}. In total, there were 36,238 green frames, 28,017 yellow frames, and 24,269 red frames. The ground truth LVEF of each subject was calculated as the mean LVEF visually assessed by a cardiac anesthesiologist.

\noindent\textbf{Landmark Verification in EchoNet}\quad
We verify the performance of our landmark detection module on the testing subset of EchoNet Dynamic by examining its ability to localize clinically significant anatomical landmarks. Figure~\ref{fig:landmark_result}A shows qualitative results, and  Figure~\ref{fig:landmark_result}B demonstrates the uncertainty in each predicted landmark. Across all visible landmarks, our method achieves an average mean Euclidean distance error (pixels) of $2.47 \pm 1.29$. Notably, for key landmarks used in the subsequent pose scoring module, we achieve error as follows:  LV apex exhibits a mean distance of $3.18 \pm 5.84$, MV $3.60 \pm 9.81$, RV $2.33 \pm 1.28$, TV $2.37 \pm 2.30$, RA $2.79 \pm 2.00$, and LA $3.06 \pm 3.69$.

\begin{figure}
    \centering
    \includegraphics[width=1\linewidth]{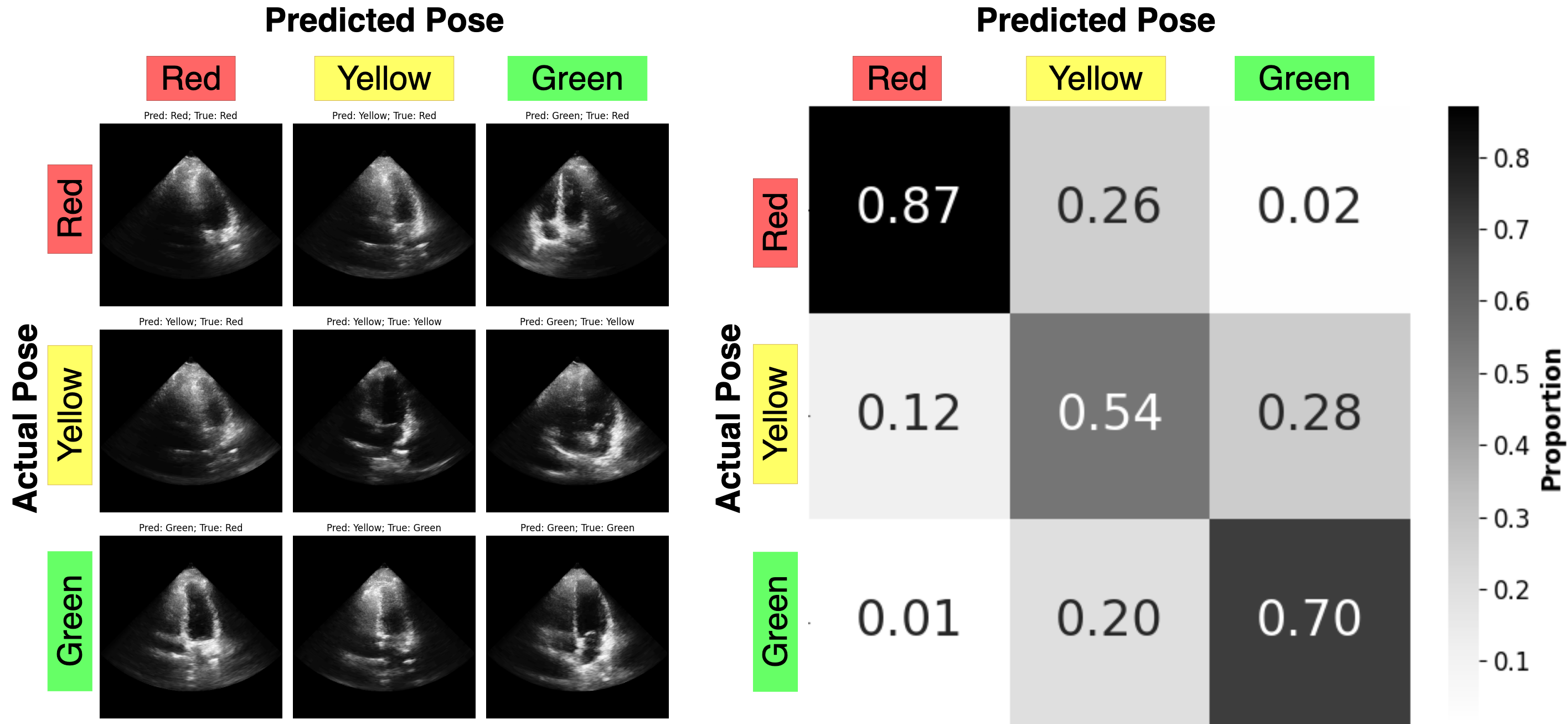}
    \caption{Overall confusion matrix and corresponding examples of misclassified TTE frames using images and landmarks. }
    \label{fig:confusion}
\end{figure}

\noindent\textbf{Pose Scoring Evaluation}\quad Overall accuracy of pose scoring is shown in Table \ref{tab_classification_accuracy_5fold}. The ResNet18 regression backbone generally has the best performance when trained with both images and landmarks. Fig.~\ref{fig:confusion}
 shows the overall confusion matrix of the ResNet18 regression backbone with examples of each type of misclassification. 
 

\noindent{\textbf{LVEF Evaluation}} The average LVEF predicted by EchoNet across 10 subjects was $50.7\% \pm 4.8\%$. The average ground truth LVEF estimated by the anesthesiologist was $59.5\% \pm 6.0\%$.  When running real-time landmark detection and pose scoring with the ResNet18 backbone, an average of 14 frames per second was achieved with 36 GB of system memory on an Apple M3 Pro chip. 


\begin{table}
\centering
\resizebox{0.49\textwidth}{!}{

\begin{tabular}{>{\centering\arraybackslash}p{15mm}>{\centering\arraybackslash}p{15mm}>{\centering\arraybackslash}p{15mm}>{\centering\arraybackslash}p{15mm}>{\centering\arraybackslash}p{15mm}>{\centering\arraybackslash}p{15mm}>{\centering\arraybackslash}p{15mm}l}
\hline
 & \multicolumn{2}{c}{Images + Landmarks} 
 & \multicolumn{2}{c}{Images Only} 
 & \multicolumn{2}{c}{Landmarks Only} & \\
\hline
 & ResNet18 \newline Regression 
 & ResNet18 \newline + LLaMA & ResNet18 \newline Regression 
 & ResNet18 \newline + LLaMA & ResNet18 \newline Regression 
 & ResNet18 \newline + LLaMA  & \\
\hline
Fold 1& 0.71& \textbf{0.73}& 0.69& 0.72& 0.64&  0.63& \\
Fold 2& 0.72& 0.72& 0.71& \textbf{0.73}& 0.64&  0.67& \\
 Fold 3& 0.68& \textbf{0.69}& 0.66& 0.67& 0.60&  0.62& \\
 Fold 4& 0.69& 0.71& 0.68& \textbf{0.73}& 0.61& 0.69& \\
 Fold 5& 0.63& 0.69& 0.68& \textbf{0.72}& 0.61& 0.65& \\
\hline
Mean & 0.69& \textbf{0.71}& 0.68& \textbf{0.71}& 0.62&  0.65& \\
\hline
\end{tabular}
}
\caption{Test accuracy of the transducer pose scoring module}
\label{tab_classification_accuracy_5fold}
\end{table}

\section{Discussion}

To our knowledge, this is the first study to provide real-time feedback on TTE transducer pose that generalizes across transducer models without relying on external tracking devices. This is achieved with a TTE “sweep” acquisition protocol that enables training of a pose scoring model across the dense anatomical region near the A4CH view. Test results demonstrate an accuracy of 71\% averaged across red, yellow, and green frames, and up to 87\% for red frames. Shown in Fig.~\ref{fig:confusion}, the main challenge is distinguishing green and yellow frames, which capture similar anatomy. Although the yellow category achieves the lowest accuracy, the results indicate that the overall approach warrants further development with larger scale data acquisition. When streamlined with automated LVEF estimation, the entire multi-task network performs real-time inference with point-of-care TTE, which may be especially valuable in physically constrained or resource-limited settings.

The automated landmarking module, in particular, has high accuracy; its incorporation of uncertainty facilitates effective dropout of landmarks as the transducer moves away from the ideal A4CH pose and provides an indication to the sonographer about image regions with weaker contrast (e.g., the lateral LV wall in Fig.~\ref{fig:landmark_result}). We observe that the accuracy of pose scoring with images and landmarks is comparable to the accuracy with images alone. Nonetheless, the combination of traffic-light feedback and anatomy-informed landmarks can provide rich visual guidance to novice users. 

With respect to architectures used for the pose scoring module, the pre-trained LLaMA backbone provides rich semantic representations, and the trainable adaption prompts efficiently transfer this knowledge to the echocardiography domain. These prompts inject domain-specific characteristics that, together with task instructions provided as prefix prompts, help guide the model toward more consistent accurate predictions.

This study has several limitations. First, although we observe stable performance in cross-validation experiments, transducer pose scoring will likely improve when trained on a larger dataset collected from diverse subjects. Secondly, pose scoring - particularly for yellow frames - is subjective despite the quantitative rubric described in Table \ref{tab:echo_scoring}.  Additionally, the method \textit{scores} the accuracy of the A4CH transducer pose within a "trafflic light" framework, but does not indicate precise transducer positioning or provide directional guidance for pose optimization, which will be the focus of future work.

\section{Conclusion}
This study demonstrates a real-time, landmark-driven TTE pose scoring system for the A4CH view that does not rely on an external transducer tracking system. The results suggest that larger-scale data acquisition and further refinement of the scoring criteria, particularly related to pose-independent factors of image quality, could yield improved accuracy. By providing automated landmark visual feedback and real-time pose scoring, the proposed approach offers a practical means to assist sonographers in resource-limited environments. 

\begingroup
\setstretch{0.8}
\bibliographystyle{IEEEtranN}
\bibliography{miccai2025}
\endgroup
\end{document}